\documentclass[letterpaper, 10 pt, conference]{ieeeconf}  
\IEEEoverridecommandlockouts                          
\title{\Large \bf
TTR-Based Reward for Reinforcement Learning with Implicit Model Priors
}
\usepackage{color}
\usepackage{bm}
\usepackage{graphics} 
\usepackage{epsfig} 
\usepackage{times} 
\usepackage{amsmath} 
\usepackage{amssymb}  
\usepackage{cite} 
\usepackage{caption}
\usepackage{subcaption}
\usepackage{float}
\usepackage{verbatim}
\usepackage{tabularx}
\usepackage{siunitx}
\usepackage{xcolor}
\usepackage{booktabs}
\usepackage[hidelinks]{hyperref}

\author{Xubo Lyu$^{1}$ and Mo Chen$^{1}$
\thanks{$^{1}$School of Computing Science, Simon Fraser University, BC, CA V5A1S6.
        {\tt\small xlv@sfu.ca},
        {\tt\small mochen@cs.sfu.ca}}%
        }

\begin{document}

\maketitle
\thispagestyle{empty}
\pagestyle{empty}

\begin{abstract}
Model-free reinforcement learning (RL) is a powerful approach for learning control policies directly from high-dimensional state and observation. 
However, it tends to be data-inefficient, which is especially costly in robotic learning tasks. 
On the other hand, optimal control does not require data if the system model is known, but cannot scale to models with high-dimensional states and observations. 
To exploit benefits of both model-free RL and optimal control, we propose time-to-reach-based (TTR-based) reward shaping, an optimal control-inspired technique to alleviate data inefficiency while retaining advantages of model-free RL. 
This is achieved by summarizing key system model information using a TTR function to greatly speed up the RL process, as shown in our simulation results. 
The TTR function is defined as the minimum time required to move from any state to the goal under assumed system dynamics constraints. 
Since the TTR function is computationally intractable for systems with high-dimensional states, we compute it for approximate, lower-dimensional system models that still captures key dynamic behaviors. 
Our approach can be flexibly and easily incorporated into any  model-free RL algorithm without altering the original algorithm structure, and is compatible with any other techniques that may facilitate the RL process.
We evaluate our approach on two representative robotic learning tasks and three well-known model-free RL algorithms, and show significant improvements in data efficiency and performance.


\end{abstract}

\section{INTRODUCTION}
Sequential decision making is a fundamental problem faced by any autonomous agent interacting extensively with environment \cite{cadd0}.
Reinforcement learning and optimal control are two essential tools for solving such problem. RL trains an agent to choose actions that maximizing its long-term accumulated reward through trial and error, and can be divided into model-free and model-based variants \cite{cadd1}. Optimal control, on the other hand, assumes the perfect knowledge of system dynamics and produces control policy through analytical computation. 

\begin{figure}[!htp]
      \centering
      \includegraphics[width=0.8\columnwidth]{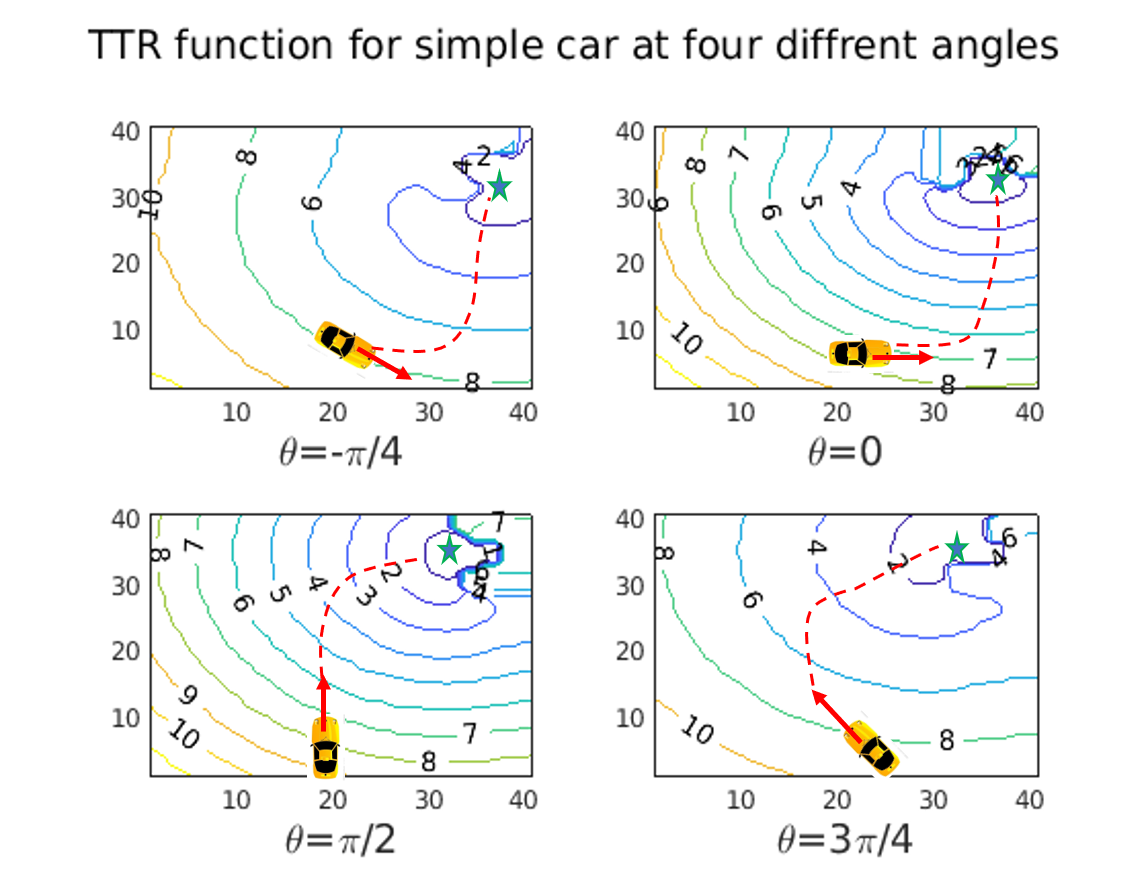}
    \caption{\small TTR functions at different heading angles for a simple car model. 
    The TTR function describes the minimum arrival time under assumed system dynamics and is effectively used for reward shaping in robotic RL tasks.}
      \label{fig:intro}
\end{figure}

Model-free RL has been successful in many fields such as games and robotics \cite{c23, c22, c45, c17, c18, c19}, and allows control policies to be learned directly from high-dimensional inputs by mapping observations to actions. 
Despite such advantages, model-free methods often require an impractically large number of trials to learn desired behaviors. 
Data inefficiency is a fundamental barrier impeding the adoption of model-free algorithms in real-world settings, especially in the context of robotics \cite{c1, c02, c44}. To address the problem of data inefficiency in model-free RL, various techniques have been proposed. ``Deep exploration" \cite{c1} samples actions from randomized value function in order to induce exploration in long term. 
Count-based exploration \cite{c40} extends near-optimal algorithms into high-dimensional state space. 
On the other hand, several recent papers refactor the structure of RL in order to utilize data more efficiently \cite{c42, c43, c44, c03, c2}. 
In particular, curriculum-based approaches \cite{c03, c2} learn progressively over multiple sub-tasks where initial task is used to guide the learner so that it will perform better on the final task. 
Hierarchical Reinforcement Learning (HRL) \cite{c42, c43, c44} involves decomposing problem into a hierarchy of sub-problems or sub-tasks such that higher-level parent-tasks invoke lower-level child tasks as if they were primitive actions.

Model-based RL uses an internal model (given or learned) that approximates the full system dynamics \cite{add1, add2, add3}. A control policy is learned based on this model. This significantly reduces the number of trials in learning and leads to fast convergence. However, model-based methods are heavily dependent on the accuracy of model itself, thus the learning performance can be easily affected by the model bias. \textcolor{black}{This is challenging especially when one aims to map sensor inputs directly to control actions, since the evolution of sensor inputs over time can be very difficult to model.}



Optimal control is an analytical method which has been substantially applied to many applications. 
For example, the authors in \cite{add4} applies optimal control on a two-joint robot manipulator in order to find robust control strategy. 
The authors in \cite{add5} realizes the real-time stabilization for a falling humanoid robot by solving a simplified optimal control problem. 
In addition, there are numerous other applications in mobile robotics and aerospace \cite{add6, add7, add8, add9}.
In general, optimal control does not require any data to generate optimal solution if the system model is known, but cannot scale to models with high-dimensional state space. 


In this paper, we propose Time-To-Reach (TTR) reward shaping, an approach that integrates optimal control into model-free RL. 
This is accomplished by incorporating into the RL algorithm a TTR-based reward function, which is obtained by solving a Hamilton-Jacobi (HJ) partial differential equation (PDE), a technique that originated in optimal control. 
A TTR function maps a robot's internal state to the minimum arrival time to the goal, assuming a model of the robot's dynamics. 
In the context of reward function in RL, intuitively a smaller TTR value indicates a desirable state for many goal-oriented robotic problems.

To accommodate the computational intractability of computing the TTR function for a high-dimensional system such as the one used in the RL problem, an approximate, low-dimensional system model that still captures key dynamic behaviors is selected for the TTR function computation. As we will demonstrate, such approximate system model is sufficient for improving data efficiency of policy learning. Therefore, our method avoids the shortcomings of both model-free RL and optimal control. 
\textcolor{black}{Unlike model-based RL, our method does not try to learn and use a full model explicitly. Instead, we maintain a looser connection between a known model and the policy improvement process in the form of a TTR reward function. This allows the policy improvement process to take advantage of model information while remaining robust to model bias.}

Our approach can be modularly incorporated into any model-free RL algorithm. 
In particular, by effectively infusing system dynamics in an implicit and compatible manner with RL, we retain the ability to learn policies that map sensor inputs directly to actions.
Our approach represents a bridge between traditional analytical optimal control approaches and the modern data-driven RL, and inherits benefits of both. 
We evaluate our approach on two common mobile robotic tasks and obtain significant improvements in learning performance and efficiency. 
We choose Proximal Policy Optimization (PPO) \cite{c22}, Trust Region Policy Optimization (TRPO) \cite{c23} and Deep Deterministic Policy Gradient (DDPG) \cite{c45} as three representative model-free algorithms to illustrate the modularity and compatibility of our approach.

\section{PRELIMINARIES}
In this section, we introduce key background  concepts of this work. 
Firstly, the Markov Decision Process is described as the fundamental mathematical framework for modeling RL problem. 
Secondly, model-free RL optimization techniques that are closely related to our work will be presented. 
Thirdly, the key concepts of approximate system model and the mathematical formulation of TTR function are given.

\subsection{Markov Decision Process} \label{sec:mdp}
A Markov Decision Process (MDP) is a discrete time stochastic control process. 
It serves as a framework for modeling decision making in situations where outcomes are partly random and partly under the control of decision maker. 
Consider a MDP defined by a 4-tuple $M=(S,A,f(\cdot,\cdot,\cdot),r(\cdot,\cdot))$, where $S$ is a finite set of states, and $A$ is a finite set of actions. $f(s,a, s')=\mathrm{Pr}(s_{t+1}=s'|s_t=s,a_t=a)$ is the transition probability that action $a$ in state $s$ at time $t$ will lead to state $s'$ at time $t+1$. 
The reward function $r(s,a)$ represents the immediate reward received if action $a$ is chosen at state $s$. 
In this paper, we employ a slight abuse of notation and write 
\begin{equation} \label{eq:MDP}
    s_{t+1} \sim f(s_t, a_t)
\end{equation} 

\noindent to denote that $s_{t+1}$ is drawn from the distribution $\mathrm{Pr}(s_{t+1}|s_t=s,a_t=a)$. 
This is done to match the notation of the approximate system dynamics presented in Eq.~\eqref{eq:approx_dyn}.


Given an MDP, one aims to find a ``policy" denoted $\pi(\cdot)$ that specifies the action $a = \pi(s)$ that is chosen at state $s$, such that the expected sum of discounted rewards
\begin{equation} \label{eq:exp_return}
    R_{\pi}(s_0) = \sum_{t=0}^{T}\gamma^t r(s_t,\pi(s_t))
\end{equation} 
\noindent is maximized over a finite horizon.
Here, $\gamma \in [0, 1]$ denotes the discount factor and is usually close to 1.


\subsection{Model-free Reinforcement Learning}
Model-free RL uses algorithms that do not require explicit knowledge of the transition probability distribution associated with the MDP to optimize RL objective. 
An obvious advantage of such algorithm is ``model-independency'' since the MDP model is often inaccessible in the problems with high-dimensional state space. 

Policy-based and value-based methods are two main approaches for training agents with model-free RL. 
Policy-based methods primarily learn a policy by representing it explicitly as $\pi_{\theta}(a|s)$ and optimizing the parameters $\theta$ either directly by gradient ascent on the performance objective $J(\pi_{\theta})$ or indirectly by maximizing local approximations of $J(\pi_{\theta})$. 
Policy-based methods sometimes involve ``on-policy" updates which means they update policy only using data collected by the most recent version of the policy. 

Value-based methods, on the other hand, primarily learn an action-value approximator $Q_{\theta}(s,a)$. 
The optimization is sometimes performed in a ``off-policy" manner which means it can learn from any trajectory sampled from the same environment. 
The corresponding policy is obtained via the connection between $Q$ and $\pi$: $\pi(s) = \arg\max_{a}Q_{\theta}(s,a)$.

Among the three model-free RL algorithms in this work, PPO and TRPO fall into the category of policy-based methods while DDPG belongs to value-based methods.

\subsection{Approximate System Model and Time-to-Reach Function}
Consider the following dynamical system in $\mathbb{R}^n$
\begin{align}
  \label{eq:approx_dyn}
  \dot{\Tilde{s}}(\tau) = \Tilde{f}(\Tilde{s}(\tau), \Tilde{a}(\tau))  
\end{align}

Note that $\Tilde{f}(\cdot)$ is used to distinguish this model from $f(\cdot)$ in Eq.~\eqref{eq:MDP}. 
Here, $\Tilde{s}(\cdot)$ and $\Tilde{a}(\cdot)$ are the state and action of an approximate system model.
The TTR problem involves finding the minimum time it takes to reach a goal from any initial state $\Tilde{s}$, subject to the system dynamics in Eq.~\eqref{eq:approx_dyn}. 
We assume that $\Tilde{f}(\cdot)$ is Lipschitz continuous. 
Under these assumptions, the dynamical system has a unique solution. 
The common approach for tackling TTR problems is to solve a Hamilton-Jacobi (HJ) partial differential equation (PDE) corresponding to system dynamics and this approach is widely applicable to both continuous and hybrid systems \cite{c25, c26, c27}.
Mathematically, the time it takes to reach a goal $\Gamma \in \mathbb{R}^n$ using a control policy $\Tilde{a}(\cdot)$ is
\begin{align}
  \label{eq:time_to_reach}
  T_{\Tilde{s}}[\Tilde{a}] = \min\{\tau|\Tilde{s}(\tau) \in \Gamma\}
\end{align}
\noindent and the TTR function is defined as follows:
\begin{align}
 \label{eq:min_ttr}
 \phi(\Tilde{s}) = \min_{\Tilde{a} \in \mathcal{\Tilde{A}}}T_{\Tilde{s}}[\Tilde{a}]
\end{align}
$\mathcal{\Tilde{A}}$ is a set of admissible controls in approximate system. Through dynamic programming, we can obtain $\phi$ by solving the following stationary HJ PDE:
\begin{align}
    \label{eq:HJPDE}
    \max_{\Tilde{a} \in \mathcal{\Tilde{A}}}\{-\nabla \phi(\Tilde{s})^{\top} \Tilde{f}(\Tilde{s},\Tilde{a}) - 1\} = 0 \\
    \phi(\Tilde{s}) = 0 ~ \forall \Tilde{s} \in \Gamma
\end{align}
Detailed derivations and discussions are presented in \cite{c31, c32}. 
Normally the computational cost of solving the TTR problem is too expensive for systems with higher than five dimensional state. 
However, model simplification and system decomposition techniques partially alleviate the computational burden in a variety of problem setups \cite{c28, c29}. 
Well-studied level set based numerical techniques \cite{c26, c27,c28,c29} have been developed to solve Eq.~\eqref{eq:HJPDE}.

\section{APPROACH}

Model-free RL algorithms have the benefit of being able to learn control policies directly from high-dimensional state and observation; however, the lack of data efficiency is a well-known challenge. 
\textcolor{black}{Integrating a fully MDP model into RL seems promising but can sometimes be difficult due to model bias. 
In this work, we address this issue by implicitly utilizing a simplified system model to provide a useful ``model-informed" reward in an important subspace of the full MDP state. This way we produce policies that are as flexible as those obtained from model-free RL algorithms, and accelerate learning without altering the model-free \mbox{pattern}.}

\begin{figure}[!htp]
    \centering
    \includegraphics[width=\columnwidth]{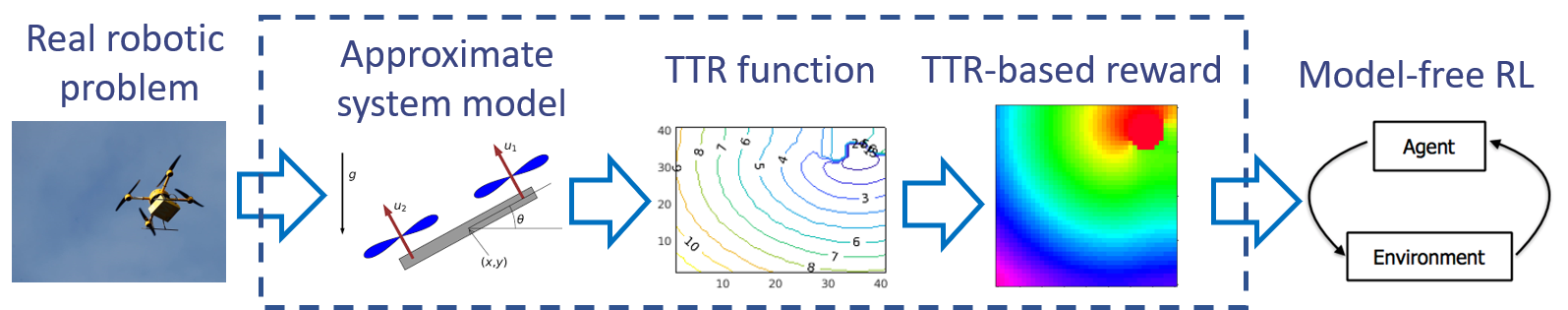}
    \caption{\small Sequential steps of TTR-based reward shaping}
    \label{fig:flow-chart}
\end{figure}

In this section, we explain the concrete steps (shown in Fig. \ref{fig:flow-chart}) of applying our method. 
The system under consideration may be represented by an MDP given by $f(\cdot)$, as explained in Section \ref{sec:mdp}; this MDP is in general unknown.
Choosing an approximate system model $\tilde f(\cdot)$ that captures key dynamic behavior is the first step; this step is explained in Section \ref{model_selection}. 
Using this approximate system model, we compute the TTR function $\phi(\cdot)$, and then apply a simple transformation to it to obtain the reward function $r(\cdot)$ that is used in RL; this is fully discussed in Section \ref{ttr_approximate_reward}. 
Finally, any model-free RL algorithm may be used to obtain a policy that maximizes the expected return in Eq.~\eqref{eq:exp_return}.

\subsection{Model Selection} \label{model_selection}
``Model selection''\footnote{Note that ``model selection'' here has a different meaning than that in machine learning.} here refers to the fact that we need to pick an (approximate) system model for the robotic task in order to compute the corresponding TTR function. 
This model should be relatively low-dimensional so that the TTR computation is tractable but still retain key behaviors in the dynamics of the system.

Before the detailed description of model selection, it is necessary to clarify some terminology used in this paper.
First, we use the phrase ``full MDP model" to refer to $f(\cdot)$, which drives the real state transitions in the RL problem. The full MDP model is often inaccessible since it captures the high-dimensional state inputs including both sensor data and robot internal state. 
Second, we will use the phrase ``approximate system model" to refer to $\Tilde{f}(\cdot)$.
The tilde indicates that $\tilde f$ does not necessarily accurately reflect the real state transitions of the problem we are solving.
In fact, the approximate system model should be low-dimensional to simplify the TTR computation while still capturing key robot physical dynamics.

The connection between the full MDP model and the approximate system model is formalized as follows. 
We assume that the approximate system state is a subset of the full MDP state. 
Thus, the relation between the full MDP state and the approximate model state is
\begin{equation}
\label{eq:state_def}
    s = (\Tilde{s}, \hat{s})
\end{equation}
Here the full state $s$ refers to the entire high-dimensional state in the full MDP model $f(\cdot)$, and $\tilde s$ refers to the state of the approximate system model $\tilde{f}(\cdot)$ which evolves according to Eq.~\eqref{eq:approx_dyn}.
For clarity, we also define $\hat s$, which are state components in the full MDP model that are not part of $\tilde s$.

For example, in the simulated car experiment in Section \ref{sec:car}, the full state contains the internal states of the car, including the position $(x,y)$, heading $\theta$, speed $v$, and turn rate $\omega$.
In addition, eight laser range measurements $d_1,\ldots,d_8$ are also part of the state $s$. These measurements provide distances from nearby obstacles.
As one can imagine, the evolution of $s$ can be very difficult if $f(\cdot)$ is impossible to obtain, especially in a  \textit{priori} unknown environments.

\begin{figure}[!htp]
    \centering
    \includegraphics[width=0.65\columnwidth]{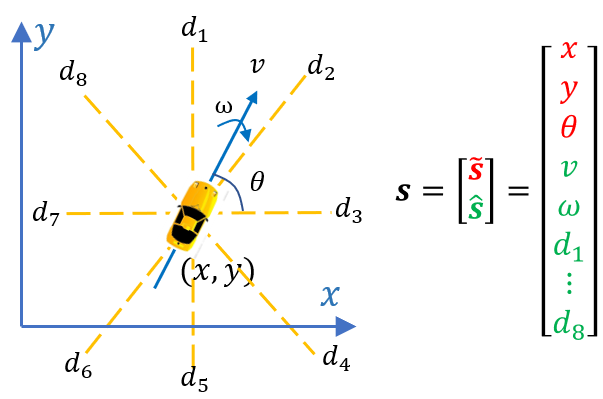}
    \caption{\small State definition of the simulated car in Section \ref{sec:car}}
    \label{fig:state_definition}
\end{figure}

The state of the approximate system, denoted $\tilde s$, contains a subset of the internal states $(x,y,\theta,v,\omega)$, and evolves according to Eq.~\eqref{eq:approx_dyn}.
In particular, for the simulated results in this paper, we choose the simple Dubins Car model to be the approximate system dynamics:
\begin{equation}
\label{eq:3d_car}
\dot{\Tilde{s}} = 
\begin{bmatrix}
\dot x\\
\dot y\\
\dot\theta\\
\end{bmatrix} =
\begin{bmatrix}
v \cos \theta \\
v \sin \theta\\
\omega \\
\end{bmatrix}
\end{equation}
As we show in Section \ref{sec:car}, such simple dynamics is sufficient for improving data efficiency in model-free RL. 
With this choice, the remaining states are denoted $\hat s = (v, \omega, d_1, \ldots, d_8)$.
Fig.~\ref{fig:state_definition} illustrates this example.
\vspace{-0.2cm}

\begin{figure*}
\begin{minipage}{0.33\linewidth}
\includegraphics[width=\textwidth]{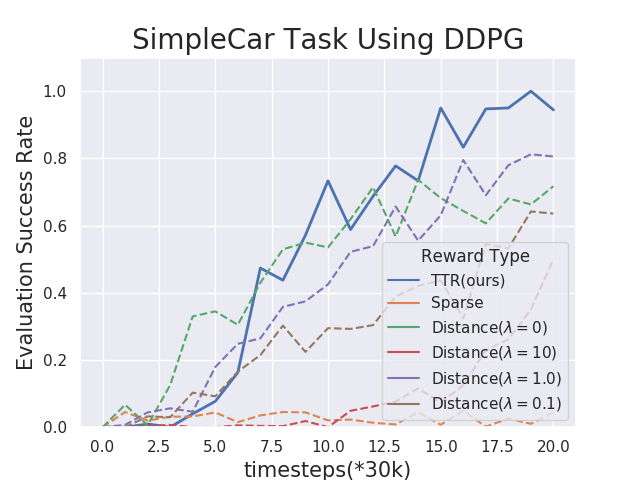}
\label{fig:sub_figure1}
\end{minipage}%
\hfill
\begin{minipage}{0.33\linewidth}
\includegraphics[width=\textwidth]{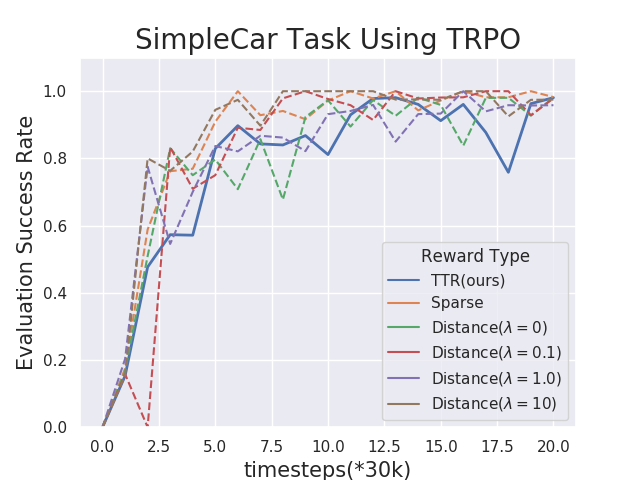}
\label{fig:sub_figure2}
\end{minipage}%
\hfill
\begin{minipage}{0.33\linewidth}
\includegraphics[width=\textwidth]{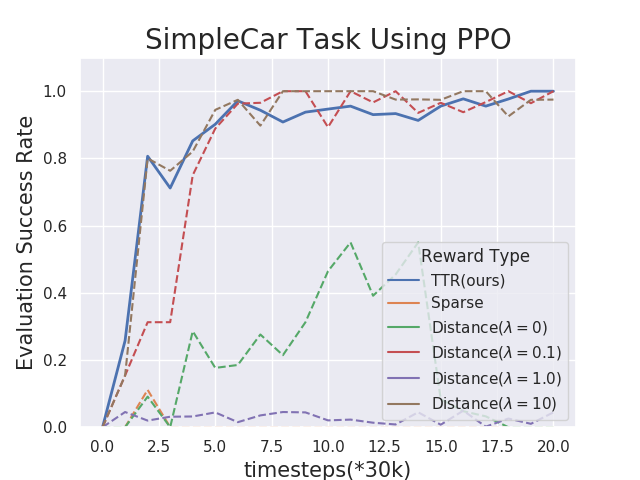}
\label{fig:sub_figure3}
\end{minipage}%
\hfill
\caption{\small Performance comparison of three different reward functions on the car example under three model-free RL optimization algorithms: DDPG, TRPO and PPO. 
All results are based on the mean of five runs. 
}
\label{fig:car_result}
\end{figure*}
In general, we may choose $\tilde s$ such that a reasonable explicit, closed-form ODE model $\Tilde{f}(\cdot)$ can be derived.
Such a model should capture the evolution of the robotic internal state. 
One motivation for using an ODE is that the real system operates in continuous time, and computing the TTR function for continuous-time systems is a solved problem for sufficiently low-dimensional systems. 

It is worth noting that if a higher-fidelity model of the car is desired, one may also choose the following 5D ODE approximate system model instead:

\begin{equation}
\label{eq:5d_car}
\dot{\Tilde{s}} = 
\begin{bmatrix}
\dot x \\
\dot y \\
\dot \theta \\
\dot v \\
\dot \omega \\
\end{bmatrix} =
\begin{bmatrix}
v \cos \theta \\
v \sin \theta \\
\omega \\
\alpha_v \\
\alpha_\omega
\end{bmatrix}
\end{equation}

In this case, we would have $\tilde s = (x,y,\theta,v,\omega)$, and $\hat s = (d_1, \ldots, d_8)$.
Note that the choice of an ODE model representing the real robot may be very flexible, depending on what behavior one wishes to capture.
In the 3D car example given in Eq.~\eqref{eq:3d_car}, we focus on modelling the position and heading of car to be consistent with the goal. However, if speed and angular speed is deemed crucial for the task under consideration, one may also choose a more complex approximate system given by Eq.~\eqref{eq:5d_car}. 
To re-iterate, a good choice of approximate model is computationally tractable for the TTR function computation, and captures the system behaviors that are important for performing the desired task.

\subsection{TTR Function as Approximate Reward} \label{ttr_approximate_reward}

In this section, we discuss how the reward function $r(s,a)$ in the full MDP can be chosen based on the TTR function. 
For simplicity, we ignore $a$ and denote it as $r(s)$, \textcolor{black}{ although a simple modification to the TTR function can be made to incorporate actions into the reward function. Since $s$ is often in high dimensional space with sensor measurements involved}, it is often unclear how to determine an proper reward for $s$. 
As a result, simple reward functions such as sparse and distance rewards are sometimes used.

However, this can be easily resolved in our approach by viewing $r(\cdot)$ as a function of $\Tilde{s}$, the state of approximate system model we have chosen before. As mentioned earlier, $\tilde{s}$ a subset of full state $s$. 
In our method, the TTR function $\phi(\Tilde{s})$ defined in Eq.~\eqref{eq:min_ttr} is transformed slightly to obtain the reward function for full MDP state $s$:
\begin{equation}
\label{eq:approx_r}
    r(s)=r(\Tilde{s}, \hat{s})=
    \begin{cases}
    -\phi(\tilde{s})&\;s\in\mathbf{I}\\
    1000& \;s\in\mathbf{G}\\
    -400& \;s\in\mathbf{C}
    \end{cases}
\end{equation}
By definition, $\phi(\Tilde{s})$ is non-negative and $\phi(\Tilde{s}) = 0$ if and only if $\Tilde{s} \in \Gamma$. 
Thus, we use $-\phi(\cdot)$ as the reward because the state with lower goal-arrival time should be given a higher reward.

\textcolor{black}{As shown in Eq.~\eqref{eq:approx_r}, we set positive reward for goal states $\mathbf{G}$ and negative reward for collision states $\mathbf{C}$. Note that the TTR-based reward can also be extended to have obstacles taken into account, or to satisfy any other design choices if required. For \emph{intermediate} states that are neither obstacles or goals, TTR function $\phi(\cdot)$ directly provides a useful reward signal in an important subspace of the full MDP state. This is significant since the associated rewards for these intermediate states are usually quite difficult to manually design, and TTR reward does not require any manual fine-tuning. This way the RL agent learns faster compared to not having a useful reward in the subspace, and can quickly learn to generalize the subspace knowledge to the high-dimensional observations. For example, positions that are near obstacles correspond to small values in LIDAR readings, and thus the agent would quickly learn these observations correspond to bad states.}



To further reduce the computational complexity of solving the PDE for more complicated system dynamics (such as a quadrotor), we may apply system decomposition methods established from the optimal control community \cite{c28,c29,c30} to obtain an approximate TTR function without significantly impacting the overall policy training time. Particularly, we first decompose the entire system into several sub-systems potentially with overlapping components of state variables, and then efficiently compute the TTR for each sub-system.  We utilize Lax-Friedrichs sweeping-based \cite{c26} to compute the TTR function. \textcolor{black}{As shown in the Table~\ref{table:TTR computation}, computation time of TTR functions are negligible compared to the time it takes to train policies.}
\section{SIMULATED EXPERIMENTS}

\begin{figure*}
\centering
\begin{subfigure}[b]{0.31\textwidth}
\includegraphics[width=\textwidth]{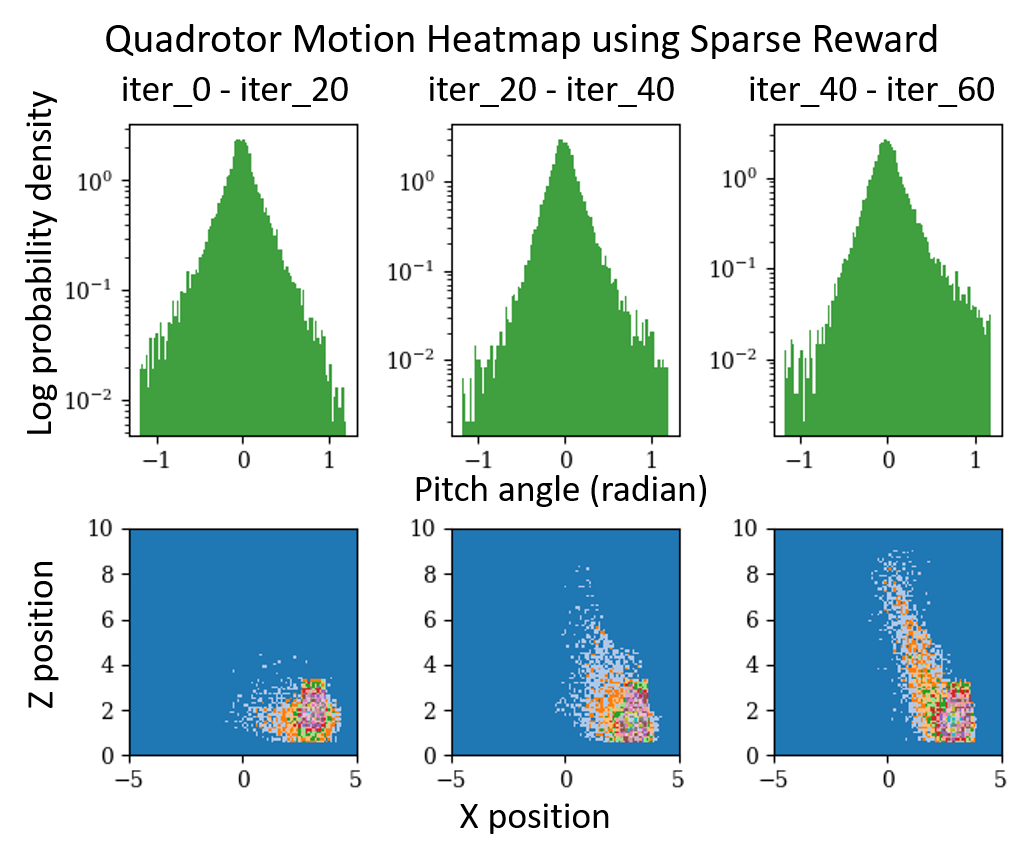}
\caption{Sparse reward}
\label{fig:heatmap1}
\end{subfigure}%
\hfill
\begin{subfigure}[b]{0.35\textwidth}
\includegraphics[width=\textwidth]{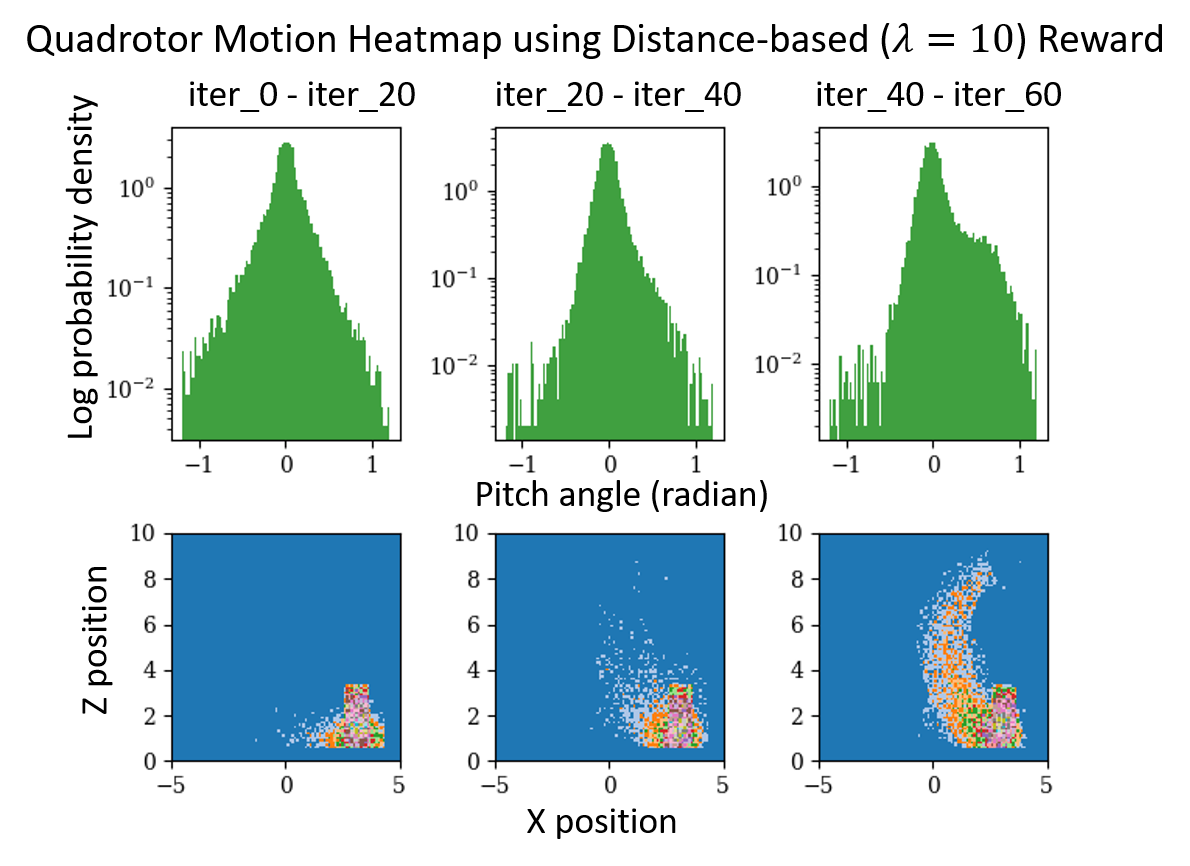}
\caption{Distance reward (the best $\lambda$)}
\label{fig:heatmap2}
\end{subfigure}%
\hfill
\begin{subfigure}[b]{0.32\textwidth}
\includegraphics[width=\textwidth]{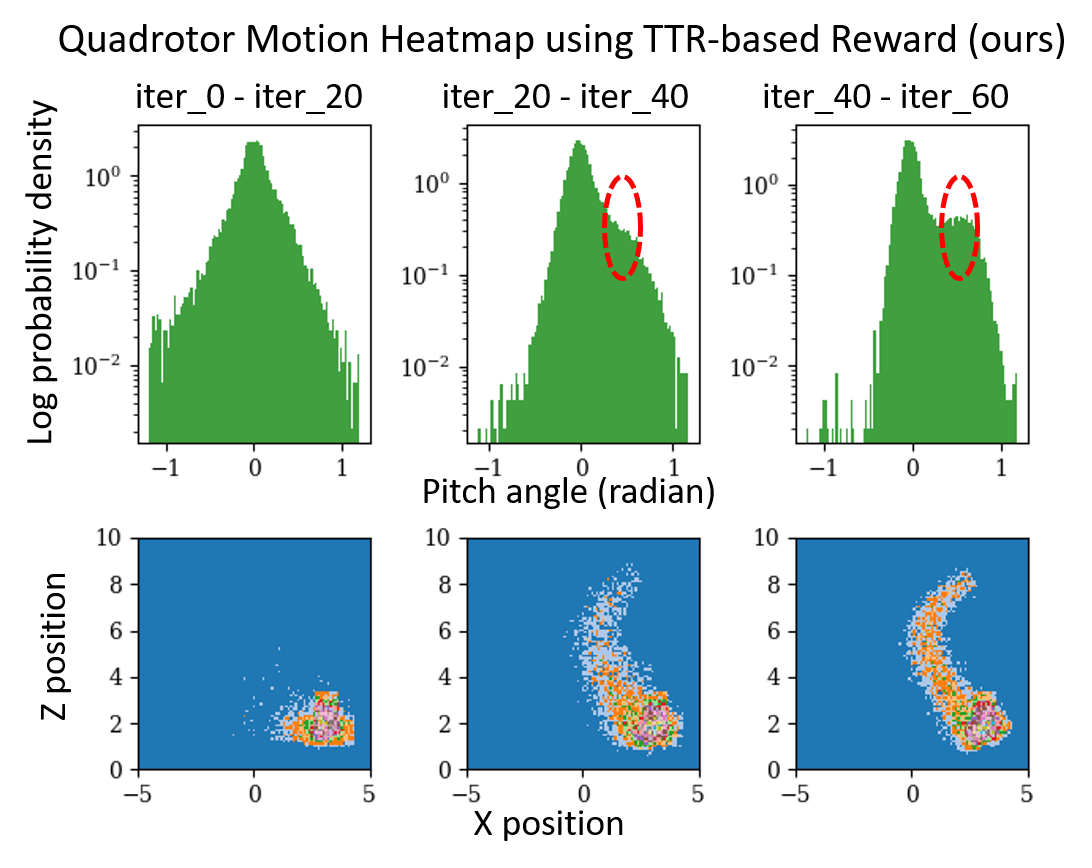}
\caption{TTR-based reward}
\label{fig:heatmap3}
\end{subfigure}%
\hfill
\caption{\small Frequency histograms of $(x,z,\psi)$ during different learning stages on quadrotor task. Top row: log probability density vs. $\psi$; bottom row: $(x,z)$ heatmap. Only the TTR-based reward leads to near-complete trajectories in the $(x,z)$ heatmap between iterations $20$ and $40$, when the other rewards still involve much exploration. Also, the shift (\textcolor{black}{\textcolor{red}{circled} in red on Fig.~\ref{fig:heatmap3}}) of log probability density towards the target $\theta$ at $0.75$ rad occurs only when TTR-based reward is used, which suggests TTR function is guiding learning effectively.}
\label{fig:heatmap}
\end{figure*}
In order to illustrate the benefits from our TTR-based reward shaping method, we now present two goal-oriented tasks through two different mobile robotic systems: a simple car and a planar quadrotor. 
Each system is simulated in Gazebo \cite{c36}, an open-source 3D physical robot simulator. 
Also, we utilize the Robot Operating System (ROS) for communication management between robot and simulator. 
For each task, we compare the performance between our TTR-based reward and two other conventional rewards: sparse and distance-based reward. 
For each reward function, we use three representative model-free RL algorithms (DDPG, TRPO and PPO) to demonstrate that TTR-based reward can be applied to augment any model-free RL algorithm. 

We select sparse and distance-based rewards for comparison with our proposed TTR-based reward because they are simple, easy to interpret, and easy to apply to any RL problem. 
These reward functions are consistent across our two simulated environments, shown in Table~\ref{table:reward functions}.
We formulate the distance-based reward as general Euclidean distance involving position and angle
because the tasks we consider involve reaching some desired set of positions and angles, shown in Table~\ref{table:reward functions}.
By choosing different weights $\lambda$, the angle is assigned different weights. 
For both examples, we choose four different weights, $\lambda \in \{0, 0.1, 1, 10\}$. 
The sparse reward is defined to be $0$ everywhere except for goal states ($1000$ reward) or collision states ($-400$ reward). 


\begin{table}[ht]
\centering
\begin{tabularx}{\columnwidth}{cccc}
\toprule
Task&Model&Computation Time&Decomposed\\
\midrule
Simple Car&Eq. \eqref{eq:3d_car}&5 sec&No\\
Planar Quadrotor&Eq. \eqref{eq:6d_quad}&90 sec&Yes\\
\bottomrule
\end{tabularx}
\caption{\small TTR function computational load. `Decomposed' means if we need to decompose approximate model into subsystems in order to reduce computational cost.}
\label{table:TTR computation}
\end{table}%

\subsection{Simple Car} \label{sec:car}

The car model is widely used as standard testbed in motion planning \cite{c36} and RL \cite{c37} tasks. 
Here we use a ``turtlebot-2" ground robot to illustrate the performance of the TTR-based reward. 
The state and observation of this example are already discussed at \ref{model_selection}.
The car starts with randomly-sampled initial conditions from the starting area and aims to reach the goal region without colliding with any obstacle along the trajectory. 
Specifically, we set the precise goal state as $G: (x_g = \SI{4}{\metre}, y_g=\SI{4}{\metre}, \theta_g=\SI{0.75}{\radian})$, and states within  $\SI{0.3}{\metre}$ in positional distance and $\SI{0.3}{\radian}$ in angular distance of $G$ are considered to have reached the goal, denoted as $S_{g} = \{(x,y,\theta)|\SI{3.7}{\metre} \leq x \leq \SI{4.3}{\metre}; \SI{3.7}{\metre} \leq y \leq \SI{4.3}{\metre}; \SI{0.45}{\radian} \leq \theta \leq \SI{1.05}{\radian} \}$. 
The TTR-based reward for this simple car task is derived from a lower-dimensional approximate car system in Eq.~\eqref{eq:3d_car} which only considers the 3D vector $(x,y, \theta)$ as state and angular velocity $\omega$ as control. 

\begin{figure*}[!htp]
\begin{minipage}{0.33\linewidth}
\includegraphics[width=\textwidth]{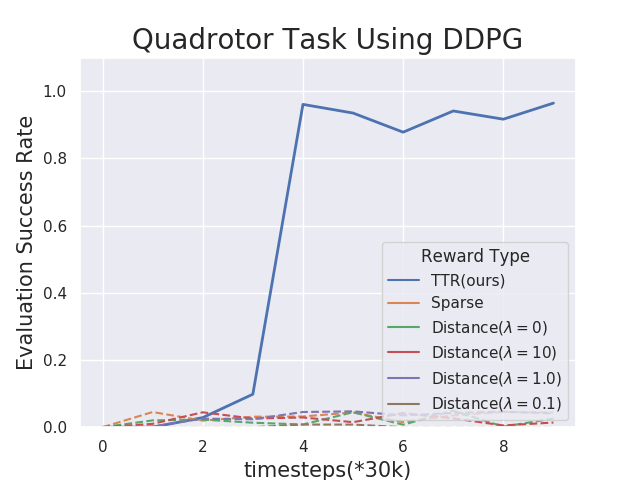}
\label{fig:subfigure1}
\end{minipage}%
\hfill
\begin{minipage}{0.33\linewidth}
\includegraphics[width=\textwidth]{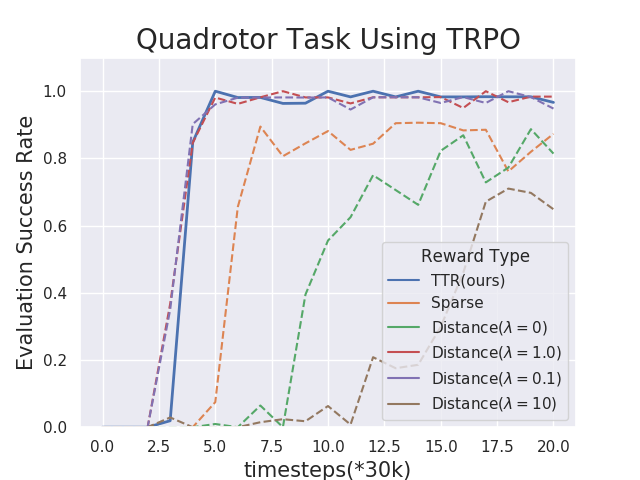}
\label{fig:subfigure2}
\end{minipage}%
\hfill
\begin{minipage}{0.33\linewidth}
\includegraphics[width=\textwidth]{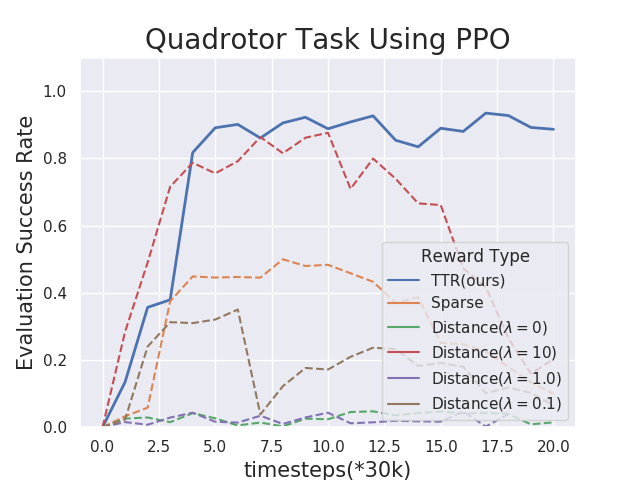}
\label{fig:subfigure3}
\end{minipage}
\hfill
\caption{\small Performance comparison between TTR, distance and sparse based rewards on quadrotor using three different model-free algorithms. The results are based on identical evaluation setting as car example and are concluded from five runs as well. Our TTR-based reward achieves the best in terms of efficiency and performance. \textbf{Left}: success rate comparison under DDPG algorithm \textbf{Middle}: success rate comparison under TRPO algorithm \textbf{Right}: success rate comparison under PPO algorithm}
\label{fig:quad_result}
\end{figure*}

Fig.~\ref{fig:car_result} compares the performance of TTR-based reward with sparse and distance-based rewards under three different model-free algorithms. 
Success rate after every fixed number of training episodes is considered as qualitative assessment. 
In general, the car system is simpler and more stable thus obtains relatively higher success rate among different reward settings compared to the quadrotor task (shown later in Fig.~\ref{fig:quad_result}). 
In particular, TTR-based reward leads to high success rate (mostly over 90\%) consistently with all three learning algorithms. 
In contrast, sparse reward leads to poor performance with PPO, and distance-based reward leads to poor performance with DDPG.

Note that despite of the better performance from certain distance-based reward, the choice of appropriate weight for each variable is non-trivial and not transferable between different tasks. 
However, TTR-based reward requires little human engineering to design and can be efficiently computed once an low-fidelity model is provided.  

The TTR function for the model in Eq.~\eqref{eq:3d_car} is shown in Fig.~\ref{fig:intro} to convey the usefulness of TTR-based reward more intuitively. 
Here, we show the 2D slices of the TTR function at four heading angles,  $\theta \in \{-\pi/4, 0, \pi/2, 3\pi/4\}$. 
The green star located at the upper-right of each plot is the goal area. 
The car starts moving from lower-middle area. 
Note that the 2D slices look different for different heading angles according to the system dynamics, with the contours expanding roughly in opposite direction to the heading slice.

\begin{table}[ht]
\setlength{\tabcolsep}{3pt}
\centering
\begin{tabularx}{\columnwidth}{cccc}
\toprule
Sparse&Distance&TTR\\
\midrule
$r(s)=
\begin{cases}
0&\\
1000&\\
-400&
\end{cases}$&$r(s)=
\begin{cases}
-d(\cdot)^*&\\
1000&\\
-400&
\end{cases}$&$r(s)=
\begin{cases}
-\phi(\tilde{s})^*&\;s\in\mathbf{I}\\
1000& \;s\in\mathbf{G}\\
-400& \;s\in\mathbf{C}
\end{cases}$\\
\midrule
\multicolumn{4}{c}{$^*d(\cdot) = \begin{cases}\sqrt{(x-x_g)^2 + (y-y_g)^2 + \lambda(\theta-\theta_g)^2}& \textbf{\scriptsize Simple Car}\\
\sqrt{(x-x_g)^2 + (z-z_g)^2 + \lambda(\psi-\psi_g)^2}& \textbf{\scriptsize Planar Quadrotor}\end{cases}$}\\
\midrule
\multicolumn{4}{l}{$^*\phi(\tilde{s})$: \text{TTR function defined in a subspace of $s$}}\\
\bottomrule
\end{tabularx}
\caption{\small Reward functions tested in this work. $\mathbf{I}$: set of intermediate states; $\mathbf{G}$: set of goal states; $\mathbf{C}$: set of collision states. $d(\cdot)$: generalized distance function involving angle.}
\label{table:reward functions}
\end{table}%

\subsection{Planar Quadrotor Model} \label{sec:quad}

A Quadrotor is usually considered difficult to control mainly because of its nonlinear and under-actuated dynamics. 
In the second experiment, we select a planar quadrotor model \cite{c51, c52}, a popular test subject in the control literature, as a relatively complex mobile robot to validate that the TTR-based reward shaping method still works well even on highly dynamic and unstable system. 
"Planar" here means the quadrotor only flies in the vertical ($x$-$z$) plane by changing the pitch angle without affecting the roll and yaw angle. 

The approximate system model has 6D internal state $\Tilde{s} = (x, v_x, z, v_z, \psi, \omega)$, where $x, z, \psi$ denote the planar positional coordinates and pitch angle, and $v_x, v_z, \omega$ denote their time derivatives respectively. 
The dynamics used for computing the TTR function are given in Eq.~\eqref{eq:6d_quad}.
The quadrotor's movement is controlled by two motor thrusts, $T_1$ and $T_2$. 
The quadrotor has mass $m$, moment of inertia $I_{yy}$, and half-length $l$. Furthermore, $g$ denotes the gravity acceleration, $C_D^v$ the translation drag coefficient, and $C_D^\psi$ the rotational drag coefficient.
Similar to the car example, the full state $s$ contains eight laser readings extracted from the ``Hokoyu\_utm30lx'' ranging sensor for detecting obstacles, in addition to the internal state $\tilde s$. 
The objective of the quadrotor is to learn a policy mapping from states and observations to thrusts that leads it to the goal region. 

\vspace{-0.3cm}
\begin{small}
\begin{equation}
\label{eq:6d_quad}
    \dot{\Tilde{s}} =
	\begin{bmatrix}
	\dot x\\
	\dot v_x\\
	\dot z\\
	\dot v_z\\
	\dot \psi\\
	\dot \omega
	\end{bmatrix}
	=
    \begin{bmatrix}
	v_x\\
	-\frac{1}{m}C^v_D v_x + \frac{T_1}{ m}\sin\psi + \frac{T_2}{m}\sin\psi \\
	v_z\\
	-\frac{1}{m}\left(mg+C^v_D v_z\right) + \frac{T_1}{ m}\cos\psi + \frac{T_2}{m}\cos\psi\\
	\omega\\
	-\frac{1}{ I_{yy}}C^\psi_D\omega + \frac{ l}{I_{yy}} T_1 - \frac{l}{I_{yy}} T_2
    \end{bmatrix}
    \hspace{-.4em}
\end{equation}  
\end{small}

\begin{figure}
      \centering
      \includegraphics[width=0.7\columnwidth]{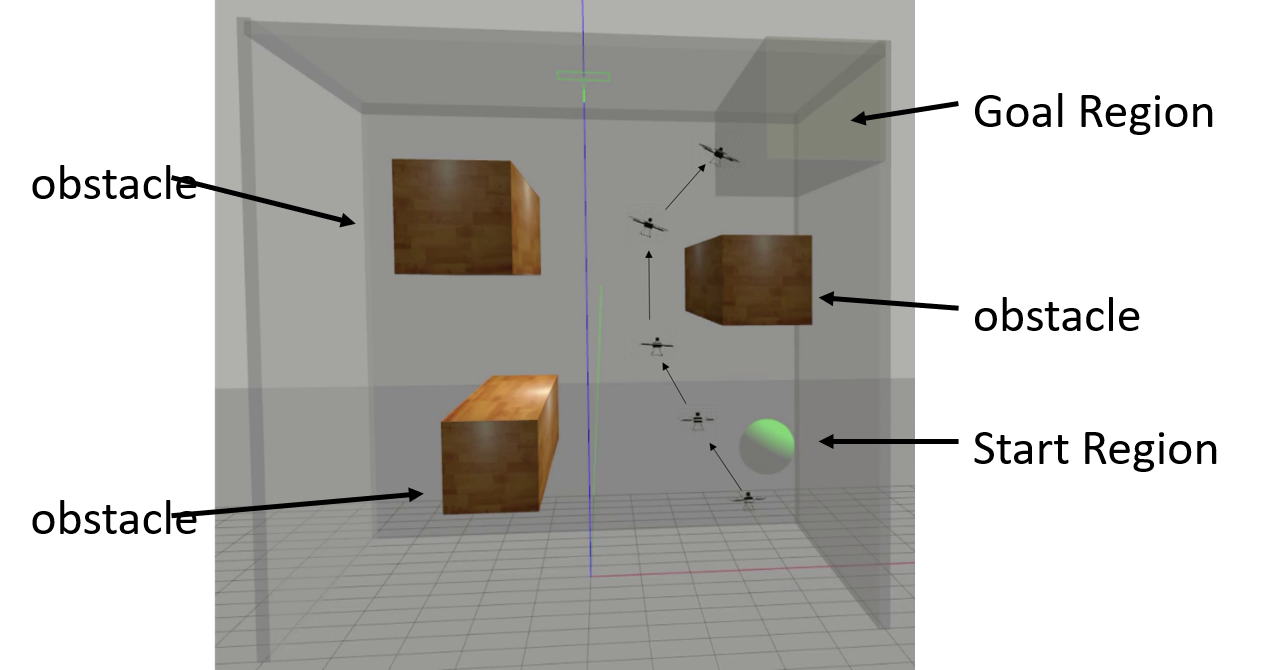}
      \caption{\small Visualization of quadrotor's sequential movement after learning from TTR-based reward. The trajectory is connected by a combination of the same quadrotor at a few different time snapshots. As shown in the picture, the quadrotor has learned to make use of  physical dynamics (tilt) to reach the target as soon as possible}
      \label{fig:gazebo_vis}
\end{figure} 

The environment for this task is shown in Fig.~\ref{fig:gazebo_vis}.  
The obstacles are fixed. 
The goal region is $S_{g} = \{(x,z, \psi)|\SI{3.5}{\metre} \leq x \leq \SI{4.5}{\metre}; \SI{8.5}{\metre} \leq z \leq \SI{9.5}{\metre}; \SI{0.45}{\radian} \leq \psi \leq \SI{1.05}{\radian}\}$. 
The quadrotor's starting condition is uniformly-randomly sampled from  $\{(x,z)|\SI{2.5}{\metre}\leq x \leq \SI{3.5}{\metre}; \SI{2.5}{\metre} \leq z \leq \SI{3.5}{\metre}\}$ (green area in Fig. \ref{fig:gazebo_vis}) and the starting pitch angle is randomly sampled from $\{\psi|\SI{-0.17}{\radian} \leq \psi \leq \SI{0.17}{\radian} \}$. 



Fig. \ref{fig:quad_result} shows the performance of TTR-based, distance-based, and sparse reward under optimization from DDPG, TRPO and PPO. 
With TTR-based reward, performance is consistent over all model-free algorithms. 
In contrast, sparse and distance-based rewards often do not lead to quadrotor stability and consistency of performance.
For example, under sparse and distance-based rewards, performance is relatively good under TRPO, but very poor under DDPG. 
In terms of learning efficiency, TTR-based reward achieves a success rate of greater than 90\% after 3 iterations (approximately 90000 time steps) regardless of the model-free algorithm. 
This is the best result among the three reward functions we tested.


To further illustrate the effectiveness of our approach, Fig.~\ref{fig:heatmap} shows statistics of positional $(x, z)$ and angular variables $\psi$ during early learning stages using the three reward functions over three ranges of learning iterations. Note that we choose distance-based reward with $\lambda=10$ since it's the best one among all tested distance-based reward variants.
The 2D histograms show the frequency of $(x, z)$ along trajectories in training episodes as heatmaps while the 1D histograms show the frequency of $\psi$ as log probability densities. 
Note that the second column of each subplot (from iteration $20 \sim 40$) represents transitory behaviors, before the quadrotor successfully learns to perform the task. 
Fig.~\ref{fig:heatmap3} shows that the desired angular goal ($\psi=0.75$ rad) has higher probability density (circled in red), which means TTR-based reward does provide effective angular local feedback. 
Furthermore, heat maps for TTR-based reward (Fig.~\ref{fig:heatmap3}) is concentrated around plausible trajectories for reaching the goal, while heat maps for the other rewards are more spread out.
This shows  TTR-based reward is providing dynamics-informed guidance.



\section{CONCLUSION}
In this paper, we propose TTR-based reward shaping to alleviate the data inefficiency of model-free RL on robotic tasks. 
By using TTR function to provide RL reward, the model-free learning process remains flexible but is endued with global guidance provided by implicit system dynamics priors.
In this approach, an approximate system model chosen in a highly flexible way. 
By computing a TTR function based on the chosen model and integrating it as the RL reward function, the agent receives more dynamics-informed feedback and learns faster and better.

Simple and effective, TTR-based reward shaping is easy to implement and can be used as a wrapper for any model-free RL algorithm since it does not alter the original algorithmic structure. 
Accordingly, any additional tricks or improvements on model-free algorithms can be attached in a compatible way. 
From the perspective of reward shaping, our approach provides a straight-forward yet distinct shaping option which requires little human engineering.

\textcolor{black}{Our method is effective when an explicit robotic system dynamics are accessible. Data efficiency can be greatly improved even if only low-dimensional approximate system dynamics are available. }


\bibliography{reference}

\begin{thebibliography}{10}
\providecommand{\url}[1]{#1}
\csname url@samestyle\endcsname
\providecommand{\newblock}{\relax}
\providecommand{\bibinfo}[2]{#2}
\providecommand{\BIBentrySTDinterwordspacing}{\spaceskip=0pt\relax}
\providecommand{\BIBentryALTinterwordstretchfactor}{4}
\providecommand{\BIBentryALTinterwordspacing}{\spaceskip=\fontdimen2\font plus
\BIBentryALTinterwordstretchfactor\fontdimen3\font minus
  \fontdimen4\font\relax}
\providecommand{\BIBforeignlanguage}[2]{{%
\expandafter\ifx\csname l@#1\endcsname\relax
\typeout{** WARNING: IEEEtran.bst: No hyphenation pattern has been}%
\typeout{** loaded for the language `#1'. Using the pattern for}%
\typeout{** the default language instead.}%
\else
\language=\csname l@#1\endcsname
\fi
#2}}
\providecommand{\BIBdecl}{\relax}
\BIBdecl

\bibitem{cadd0}
M.~L. Littman, \emph{Algorithms for sequential decision making}, 1996.

\bibitem{cadd1}
A.~S. Polydoros and L.~Nalpantidis, ``Survey of model-based reinforcement
  learning: Applications on robotics,'' \emph{J. Intelligent {\&} Robotic
  Systems}, vol.~86, no.~2, pp. 153--173, May 2017.

\bibitem{c23}
J.~Schulman, F.~Wolski, P.~Dhariwal, A.~Radford, and O.~Klimov, ``Proximal
  policy optimization algorithms,'' \emph{CoRR}, vol. abs/1707.06347, 2017.

\bibitem{c22}
J.~Schulman, S.~Levine, P.~Moritz, M.~Jordan, and P.~Abbeel, ``Trust region
  policy optimization,'' in \emph{Proc. Annual Int. Conf. Machine Learning},
  2015.

\bibitem{c45}
T.~P. Lillicrap, J.~J. Hunt, A.~Pritzel, N.~Heess, T.~Erez, Y.~Tassa,
  D.~Silver, and D.~Wierstra, ``Continuous control with deep reinforcement
  learning,'' \emph{arXiv preprint arXiv:1509.02971}, 2015.

\bibitem{c17}
R.~S. Sutton and A.~G. Barto, \emph{Reinforcement Learning: An
  Introduction}.\hskip 1em plus 0.5em minus 0.4em\relax USA: A Bradford Book,
  2018.

\bibitem{c18}
J.~Schmidhuber, ``Deep learning in neural networks: An overview,'' \emph{Neural
  Networks}, vol.~61, pp. 85 -- 117, 2015.

\bibitem{c19}
Y.~{Lecun}, Y.~{Bengio}, and G.~{Hinton}, ``{Deep learning},'' \emph{Nature},
  vol. 521, pp. 436--444, May 2015.

\bibitem{c1}
I.~Osband, B.~Van~Roy, D.~Russo, and Z.~Wen, ``Deep exploration via randomized
  value functions,'' \emph{arXiv preprint arXiv:1703.07608}, 2017.

\bibitem{c02}
A.~Guez, D.~Silver, and P.~Dayan, ``Efficient bayes-adaptive reinforcement
  learning using sample-based search,'' in \emph{Advances in Neural Information
  Processing Systems 25}, F.~Pereira, C.~J.~C. Burges, L.~Bottou, and K.~Q.
  Weinberger, Eds.\hskip 1em plus 0.5em minus 0.4em\relax Curran Associates,
  Inc., 2012, pp. 1025--1033.

\bibitem{c44}
O.~Nachum, S.~S. Gu, H.~Lee, and S.~Levine, ``Data-efficient hierarchical
  reinforcement learning,'' in \emph{Advances in Neural Information Processing
  Systems}, 2018, pp. 3303--3313.

\bibitem{c40}
H.~Tang, R.~Houthooft, D.~Foote, A.~Stooke, O.~X. Chen, Y.~Duan, J.~Schulman,
  F.~DeTurck, and P.~Abbeel, ``\#exploration: A study of count-based
  exploration for deep reinforcement learning,'' in \emph{Advances in neural
  information processing systems}, 2017, pp. 2753--2762.

\bibitem{c42}
A.~Vezhnevets, V.~Mnih, S.~Osindero, A.~Graves, O.~Vinyals, J.~Agapiou
  \emph{et~al.}, ``Strategic attentive writer for learning macro-actions,'' in
  \emph{Advances in neural information processing systems}, 2016, pp.
  3486--3494.

\bibitem{c43}
A.~S. Vezhnevets, S.~Osindero, T.~Schaul, N.~Heess, M.~Jaderberg, D.~Silver,
  and K.~Kavukcuoglu, ``Feudal networks for hierarchical reinforcement
  learning,'' in \emph{Proceedings of the 34th International Conference on
  Machine Learning-Volume 70}.\hskip 1em plus 0.5em minus 0.4em\relax JMLR.
  org, 2017, pp. 3540--3549.

\bibitem{c03}
Y.~Bengio, J.~Louradour, R.~Collobert, and J.~Weston, ``Curriculum learning,''
  in \emph{Proc. Annual Int. Conf. Machine Learning}, 2009.

\bibitem{c2}
\BIBentryALTinterwordspacing
C.~Florensa, D.~Held, M.~Wulfmeier, and P.~Abbeel, ``Reverse curriculum
  generation for reinforcement learning,'' \emph{CoRR}, 2017. [Online].
  Available: \url{http://arxiv.org/abs/1707.05300}
\BIBentrySTDinterwordspacing

\bibitem{add1}
C.~G. Atkeson, A.~W. Moore, and S.~Schaal, ``Locally weighted learning for
  control,'' in \emph{Lazy learning}.\hskip 1em plus 0.5em minus 0.4em\relax
  Springer, 1997, pp. 75--113.

\bibitem{add2}
P.~Abbeel, A.~Coates, M.~Quigley, and A.~Y. Ng, ``An application of
  reinforcement learning to aerobatic helicopter flight,'' in \emph{Advances in
  neural information processing systems}, 2007, pp. 1--8.

\bibitem{add3}
M.~Deisenroth and C.~E. Rasmussen, ``Pilco: A model-based and data-efficient
  approach to policy search,'' in \emph{Proceedings of the 28th International
  Conference on machine learning (ICML-11)}, 2011, pp. 465--472.

\bibitem{add4}
F.~Lin and R.~D. Brandt, ``An optimal control approach to robust control of
  robot manipulators,'' \emph{IEEE Transactions on Robotics and Automation},
  vol.~14, no.~1, pp. 69--77, 1998.

\bibitem{add5}
S.~{Wang} and K.~{Hauser}, ``Realization of a real-time optimal control
  strategy to stabilize a falling humanoid robot with hand contact,'' in
  \emph{2018 IEEE International Conference on Robotics and Automation (ICRA)},
  May 2018, pp. 3092--3098.

\bibitem{add6}
M.~Chen and C.~J. Tomlin, ``Hamilton--jacobi reachability: Some recent
  theoretical advances and applications in unmanned airspace management,''
  \emph{Annual Review of Control, Robotics, and Autonomous Systems}, vol.~1,
  pp. 333--358, 2018.

\bibitem{add7}
M.~Chen, Q.~Hu, J.~F. Fisac, K.~Akametalu, C.~Mackin, and C.~J. Tomlin,
  ``Reachability-based safety and goal satisfaction of unmanned aerial platoons
  on air highways,'' \emph{Journal of Guidance, Control, and Dynamics},
  vol.~40, no.~6, pp. 1360--1373, 2017.

\bibitem{add8}
M.~Chen, J.~F. Fisac, S.~Sastry, and C.~J. Tomlin, ``Safe sequential path
  planning of multi-vehicle systems via double-obstacle hamilton-jacobi-isaacs
  variational inequality,'' in \emph{2015 European Control Conference
  (ECC)}.\hskip 1em plus 0.5em minus 0.4em\relax IEEE, 2015, pp. 3304--3309.

\bibitem{add9}
M.~Chen, J.~C. Shih, and C.~J. Tomlin, ``Multi-vehicle collision avoidance via
  hamilton-jacobi reachability and mixed integer programming,'' in \emph{2016
  IEEE 55th Conference on Decision and Control (CDC)}.\hskip 1em plus 0.5em
  minus 0.4em\relax IEEE, 2016, pp. 1695--1700.

\bibitem{c25}
Z.~{Zhou}, R.~{Takei}, H.~{Huang}, and C.~J. {Tomlin}, ``A general, open-loop
  formulation for reach-avoid games,'' in \emph{Proc. IEEE Conf, Decision and
  Control}, 2012.

\bibitem{c26}
I.~Yang, S.~Becker-Weimann, M.~J. Bissell, and C.~J. Tomlin, ``One-shot
  computation of reachable sets for differential games,'' in \emph{Proc. ACM
  Int. Conf. Hybrid Systems: Computation and Control}, 2013.

\bibitem{c27}
R.~Takei and R.~Tsai, ``Optimal trajectories of curvature constrained motion in
  the hamilton--jacobi formulation,'' \emph{J. Scientific Computing}, vol.~54,
  no.~2, pp. 622--644, Feb 2013.

\bibitem{c31}
M.~Bardi and I.~Capuzzo-Dolcetta, \emph{Optimal Control and Viscosity Solutions
  of Hamilton-Jacobi-Bellman Equations}, ser. Modern Birkh{\"a}user Classics,
  2008.

\bibitem{c32}
M.~Bardi and P.~Soravia, ``Hamilton-jacobi equations with singular boundary
  conditions on a free boundary and applications to differential games,''
  \emph{Transactions of the American Mathematical Society}, vol. 325, no.~1,
  pp. 205--229, 1991.

\bibitem{c28}
I.~M. Mitchell, ``The flexible, extensible and efficient toolbox of level set
  methods,'' \emph{J. Scientific Computing}, vol.~35, no.~2, pp. 300--329, Jun
  2008.

\bibitem{c29}
M.~{Chen}, S.~{Herbert}, and C.~J. {Tomlin}, ``Fast reachable set
  approximations via state decoupling disturbances,'' in \emph{Proc. IEEE Conf.
  Decision and Control}, 2016.

\bibitem{c30}
M.~{Chen}, S.~L. {Herbert}, M.~S. {Vashishtha}, S.~{Bansal}, and C.~J.
  {Tomlin}, ``Decomposition of reachable sets and tubes for a class of
  nonlinear systems,'' \emph{IEEE Transactions on Automatic Control}, vol.~63,
  no.~11, pp. 3675--3688, Nov 2018.

\bibitem{c36}
N.~{Koenig} and A.~{Howard}, ``Design and use paradigms for gazebo, an
  open-source multi-robot simulator,'' in \emph{Proc. IEEE/RSJ Int. Conf.
  Intelligent Robots and Systems}, 2004.

\bibitem{c37}
D.~J. {Webb} and J.~{van den Berg}, ``Kinodynamic rrt*: Asymptotically optimal
  motion planning for robots with linear dynamics,'' in \emph{Proc. IEEE
  Int.Conf. Robotics and Automation}, 2013.

\bibitem{c51}
J.~H. Gillula, H.~Huang, M.~P. Vitus, and C.~J. Tomlin, ``Design of guaranteed
  safe maneuvers using reachable sets: Autonomous quadrotor aerobatics in
  theory and practice,'' in \emph{2010 IEEE International Conference on
  Robotics and Automation}.\hskip 1em plus 0.5em minus 0.4em\relax IEEE, 2010,
  pp. 1649--1654.

\bibitem{c52}
S.~Singh, A.~Majumdar, J.-J. Slotine, and M.~Pavone, ``Robust online motion
  planning via contraction theory and convex optimization,'' in \emph{2017 IEEE
  International Conference on Robotics and Automation (ICRA)}.\hskip 1em plus
  0.5em minus 0.4em\relax IEEE, 2017, pp. 5883--5890.

\end{thebibliography}
\bibliographystyle{IEEEtran.bst}
\end{document}